\documentclass[conference]{IEEEtran}
\IEEEoverridecommandlockouts
\usepackage{cite}
\usepackage{amsmath,amssymb,amsfonts}
\usepackage{algorithmic}
\usepackage{graphicx}
\usepackage{textcomp}
\usepackage{xcolor}
\def\BibTeX{{\rm B\kern-.05em{\sc i\kern-.025em b}\kern-.08em
    T\kern-.1667em\lower.7ex\hbox{E}\kern-.125emX}}

\usepackage{booktabs}
\usepackage{multirow}
\usepackage[table]{xcolor}
\usepackage[compatibility=false]{caption}
\usepackage{subcaption}
\begin{document}

\title{PARAN: Persona-Augmented Review ANswering system on Food Delivery Review Dataset\\

% \thanks{Identify applicable funding agency here. If none, delete this. lemong}
}

% \author{
% \IEEEauthorblockN{Anonymous Author}
% }

\author{
\IEEEauthorblockN{Moonsoo Park}
\IEEEauthorblockA{\textit{Lemong Research} \\
Seoul, Republic of Korea \\
kd.mpark10@gmail.com}
\and

\IEEEauthorblockN{Jeongseok Yun}
\IEEEauthorblockA{\textit{Lemong Research} \\
Seoul, Republic of Korea\\
jeongseok.yun@lemong.ai}
\and

\IEEEauthorblockN{Bohyung Kim}
\IEEEauthorblockA{\textit{Lemong Research} \\
Seoul, Republic of Korea\\
bohyung@lemong.ai}
}

\maketitle

% \begin{abstract}
% This document is a model and instructions for \LaTeX.
% This and the IEEEtran.cls file define the components of your paper [title, text, heads, etc.]. *CRITICAL: Do Not Use Symbols, Special Characters, Footnotes, 
% or Math in Paper Title or Abstract.
% \end{abstract}
\begin{abstract}
Personalized review response generation presents a significant challenge in domains where user information is limited, such as food delivery platforms. While large language models (LLMs) offer powerful text generation capabilities, they often produce generic responses when lacking contextual user data, reducing engagement and effectiveness. In this work, we propose a two-stage prompting framework that infers both explicit (e.g., user-stated preferences) and implicit (e.g., demographic or stylistic cues) personas directly from short review texts. These inferred persona attributes are then incorporated into the response generation prompt to produce user-tailored replies. To encourage diverse yet faithful generations, we adjust decoding temperature during inference. We evaluate our method using a real-world dataset collected from a Korean food delivery app, and assess its impact on precision, diversity, and semantic consistency. Our findings highlight the effectiveness of persona-augmented prompting in enhancing the relevance and personalization of automated responses without requiring model fine-tuning.
\end{abstract}


\section{Introduction}

In real-world applications, it is often difficult to obtain sufficient background information or contextual signals about users. This challenge is particularly salient in online platforms such as food delivery apps, where interactions with users are limited. As a result, providing appropriate and personalized responses to user reviews becomes difficult, and manually responding to every review is time-consuming and costly. While many prior approaches rely on structured user metadata or historical interaction logs, our framework operates under a more realistic constraint—inferring user personas solely from short, sparse review texts, without access to any auxiliary user information.

With the recent advancement of large language models (LLMs), automated text generation systems are increasingly being deployed in a wide range of domains \cite{auto_review_answering, social_comment, chatbot}, including review response generation, social media comment automation, and counseling chatbots. However, in scenarios with limited user information, LLMs tend to produce generic responses across users \cite{generic_response1, generic_response2, generic_response3}, which can negatively affect user engagement and service satisfaction.

Previous studies \cite{auto_review_answering, auto_review_answering2} have shown that timely and relevant responses to user reviews can positively influence customer satisfaction and even drive sales. More importantly, incorporating user-specific traits or preferences—such as whether a customer prioritizes delivery speed versus food quantity—can lead to responses that better resonate with individual users \cite{personalized_review_good}.

In this work, we collect user review data from a Korean food delivery platform and investigate whether LLMs can infer both explicit persona factors (e.g., stated preferences) and implicit attributes (e.g., age, gender) from short review texts, and use these inferences to generate personalized responses. Without relying on model fine-tuning, we design a two-stage prompting strategy: the LLM first infers a likely explicit and implicit persona from the review, which is then incorporated into the final response generation prompt.

One of the key challenges is that food delivery reviews are typically short and sparse, offering limited cues for persona inference. To mitigate this limitation, we adjust the temperature parameter during inference to encourage the LLM to leverage its world knowledge and generate more diverse responses. We empirically evaluate how this approach affects the precision, diversity, and answer consistency of the generated responses. To capture these objectives, we measure precision with n-gram overlap metrics (Rouge-2, BLEU, METEOR), diversity with lexical variation (Distinct-2), and answer consistency with embedding-based semantic similarity (BERTScore).
\section{Related Work}

\begin{figure*}[!htbp]
  \centering
  \includegraphics[width = 7in]{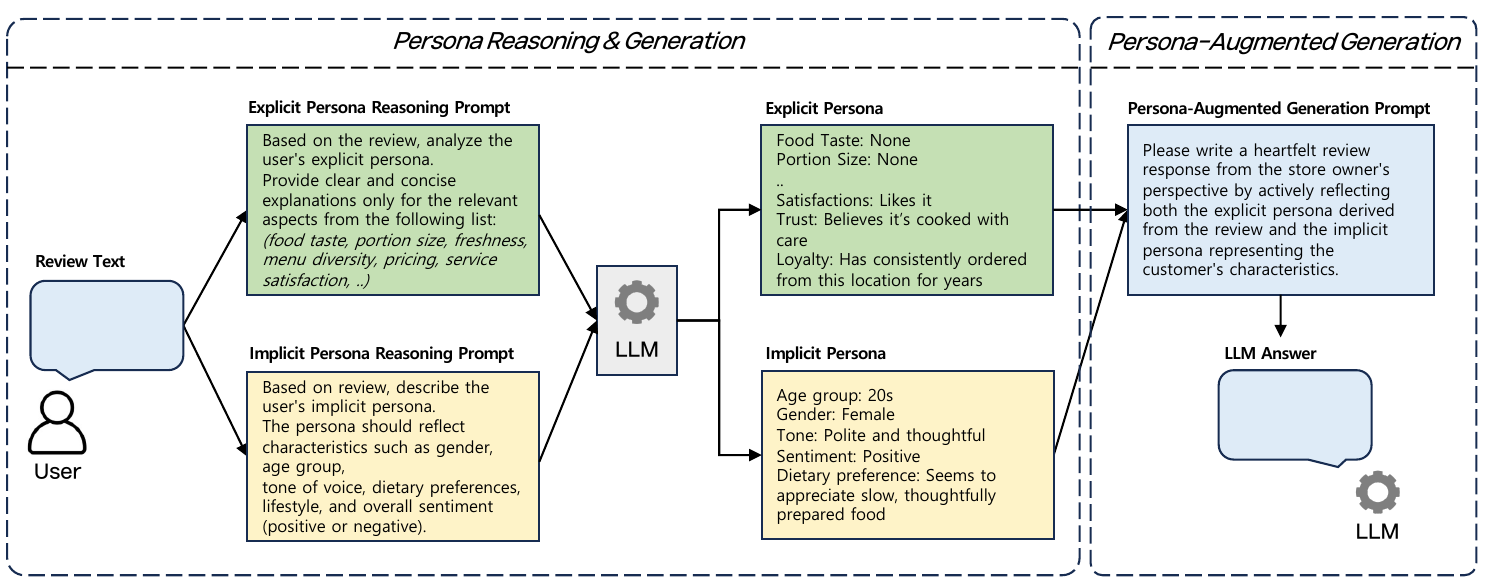}
  \caption{The overall framework of PARAN.}
  \label{fig:framework}
\end{figure*}

\subsection{Persona Integration in Language Models}
A growing body of research has explored the integration of persona information into large language models (LLMs) to enhance text generation. Early approaches often relied on manually constructed persona profiles (e.g., a set of predefined traits or background sentences), which were provided to the model as additional input to guide generation, particularly in dialogue systems \cite{zhang2018personalizing}. More recent work leverages prompt engineering or fine-tuning techniques to condition generation on persona representations with personal traits \cite{tseng2024two,liu2020you}.
% \cite{liu2021pretraining,jiang2022prompting}

Unlike these studies, our work does not assume any prior persona information and instead infers persona traits directly from the user’s input (i.e., review text) using zero-shot prompting. This allows us to explore the extent to which pretrained LLMs can infer and reflect user persona without additional training, which is critical in real-world settings where user metadata is scarce.

\subsection{Persona Inference from Dialogue or Text}
Another relevant line of research involves inferring persona characteristics—such as age, gender, sentiment, or personal preferences—from textual data. Prior work has employed classification models trained on annotated corpora \cite{mazare2018training}, probabilistic inference based on dialogue history \cite{joshi2017personalization}, or few-shot prompting with LLMs \cite{cheng2024evolving} to identify persona attributes.

In contrast, our approach focuses on zero-shot persona inference from single-shot user reviews, which are often short and sparse. Rather than relying on supervised signals or dialogue context, we prompt LLMs to infer personas on-the-fly and measure how well these inferred traits are reflected in the final responses while preserving the original text. This setup poses a more challenging but practically relevant scenario, especially for high-turnover, low-context domains like food delivery platforms.

\subsection{Review Response Generation and Personalization}
Review response generation has gained attention in domains such as e-commerce, hospitality, and app platforms \cite{zhao2019review,kew2020benchmarking}. Most existing methods focus on generating fluent and sentiment-aligned responses, with limited attention to user personalization. Some recent studies incorporate features such as user sentiment, review type, or star rating to guide generation \cite{gao2019automating}, but these are often derived from structured metadata or require domain-specific supervision.

Our work differs in two key ways: (1) we aim to personalize responses based on latent user traits inferred from unstructured review text, and (2) we do so in a model-agnostic manner without fine-tuning, relying solely on prompting strategies. This makes our approach lightweight, adaptable across models, and potentially more scalable in commercial applications.

\section{Methodology}

% \begin{figure*}[!htbp]
%   \centering
%   \includegraphics[width = 7in]{fig/figs.pdf}
%   \caption{The overall framework of PARAN.}
%   \label{fig:framework}
% \end{figure*}

\subsection{Persona Reasoning and Generation}

To generate responses that are both faithful to the original review and tailored to individual users, we model two complementary types of personas: Explicit Persona and Implicit Persona. This dual modeling approach is designed to mitigate the trade-off between content fidelity and response diversity, a common challenge in personalized generation. Relying solely on explicit review information often limits the variability of responses, while focusing only on personalization risks drifting from the original review context. By incorporating both persona types, our framework seeks to generate contextually grounded yet user-specific responses.

\noindent \textbf{Explicit Persona}. 
The Explicit Persona captures information explicitly stated in the original review text. Drawing from prior consumer behavior studies in the food domain \cite{food_factors}, we identify a set of key decision factors (e.g., taste, portion size, price, freshness, service quality) that commonly influence dining experiences. A team of human annotators verified and refined these factors to ensure reliability and coverage. Each factor is then extracted from the review by prompting LLMs to extract relevant information.

In total, the explicit persona spans 14 attributes, including but not limited to: food taste, portion size, freshness, menu diversity, price, service satisfaction, delivery experience, trust, friendliness, references to prior reviews, rating alignment, cleanliness, loyalty (e.g., returning customers), social context (e.g., companions), and temporal events (e.g., holidays, birthdays). These attributes serve to anchor the generated response in the factual content of the review and promote empathetic alignment with the reviewer.

\noindent \textbf{Implicit Persona}. 
The Implicit Persona encodes reviewer-specific traits that are not directly observable from the text but can be inferred through linguistic and stylistic patterns. These latent attributes are automatically derived by the LLM without explicit supervision, leveraging its pretraining on large-scale human interaction data.

The inferred traits include, but are not limited to: gender, age group, tone of voice, dietary preferences, lifestyle, and the reviewer’s overall sentiment polarity. While such information is not explicitly present in the review, the model is prompted to associate certain expressions, phrase structures, or lexical choices with underlying user characteristics based on its pretrained knowledge.

By incorporating the implicit persona, the system promotes personalized variation in generated responses while reducing generic or overly templated outputs. This enhances diversity and user alignment without compromising on coherence or fluency.

\subsection{Persona-Augmented Generation}

Given a review context $C$, an explicit persona $P_e$, and an implicit persona $P_i$, our goal is to generate a response $r$ that is faithful to the original review while reflecting inferred user characteristics.

We define the generation process as:

\begin{equation}
P(r \mid C, P_e, P_i, \tau; \theta) = \prod_{t=1}^{T} P(r_t \mid r_{<t}, C, P_e, P_i, \tau)
\end{equation}

Here, $\tau$ denotes the temperature parameter that controls the diversity of the generated output. A higher $\tau$ encourages more exploratory responses, especially in the presence of ambiguous or sparse persona cues.

This formulation allows us to condition generation on both structured persona ($P_e$) and inferred latent traits ($P_i$), and to systematically evaluate the trade-off between precision and diversity through temperature scaling.

\section{Experimental Setting}

\subsection{Data Collection}

% \begin{table}[!htbp]
%     \caption{Statistics of the experimented dataset.}
%     \centering
%     \begin{tabular}{l|ccc}
%         \toprule[1pt]
%          \textbf{Dataset}      & \textbf{\# Users} & \textbf{\# Items} & \textbf{\# Reviews}   \\
%          \midrule
%          \textbf{Baemin}       & 82                & 96                & 1,110                 \\
%         \bottomrule[1pt]
%     \end{tabular}
%     \label{tab:dataset}
% \end{table}

\noindent \textbf{Food Delivery Review Dataset}.
For this study, we collected user review data from Baemin, one of the largest food delivery platforms in South Korea. The dataset comprises diverse and richly contextualized user-written reviews that reflect a wide range of delivery experiences across various food categories, merchant types, geographic regions, and seasonal contexts. These reviews naturally encode users’ preferences, priorities, and attitudes, making the dataset well-suited for building personalized response generation systems.

The reviews were collected between April 2024 and April 2025, and all were originally written in Korean by real users. Although all experiments were conducted in Korean, the dataset, prompts, and model outputs were translated into English for the convenience of non-Korean speakers.

This dataset enables the generation of personalized responses grounded in user reviews, allowing us to evaluate the extent to which LLM-generated replies align with the original content in terms of empathy and personalization.

% version for author name
\begin{table}[!htbp]
    \caption{Statistics of the experimented dataset.}
    \centering
    \begin{tabular}{l|ccc}
        \toprule[1pt]
         \textbf{Dataset}      & \textbf{\# Users} & \textbf{\# Items} & \textbf{\# Reviews}   \\
         \midrule
         \textbf{Baemin}       & 82                & 96                & 1,110                 \\
        \bottomrule[1pt]
    \end{tabular}
    \label{tab:dataset}
\end{table}

\noindent \textbf{Data Preprocessing}.
To ensure the textual input contained sufficient information for persona inference, we removed reviews consisting of five or fewer words, following minimum-length filtering practices commonly used in dialogue generation literature \cite{fiveword}.

To distinguish between different users who share the same nickname, we redefined reviewer identifiers by combining the nickname with the address of the merchant at the time of review submission. For example, if the same nickname appeared in reviews targeting restaurants located in different districts, we treated them as separate users.

Despite this disambiguation, we observed that some accounts exhibited abnormally high levels of review activity. We therefore excluded users who wrote more than 365 reviews in a year, regarded them as outliers \cite{mukherjee2013spotting}. Finally, to retain only users and merchants with sufficient interaction history for meaningful persona modeling, we applied k-core filtering with k = 10 on both reviewers and merchants. Please refer to Table \ref{tab:dataset} for the dataset statistics.

\subsection{Evaluation Metrics}

To evaluate the lexical fidelity between the generated response and the original review, we report Rouge-2 \cite{rouge}, BLEU \cite{bleu}, and METEOR \cite{meteor}. These token-level overlap metrics are widely used to assess generation precision and content alignment in text generation tasks. To measure lexical diversity, we report Distinct-2 \cite{distinct}, calculated as the number of distinct bigrams divided by the total number of generated tokens. While not a direct measure of personalization, higher diversity can indicate the model's ability to produce user-specific, non-generic responses under varying persona conditions.

\subsection{LLMs}

To evaluate the effect of persona integration on response generation, we selected a diverse set of both open-source and closed-source LLMs. For closed-source models, we included GPT-4o mini, GPT-3.5 Turbo, Claude 3.5 Haiku, and Amazon Nova Lite, which represent commercially deployed systems with varying architectures and capabilities. For open-source models, we used Llama 3.1 Instruct (8B) and Llama 3.1 Instruct (70B) from Meta. This selection facilitates a broad comparison of persona-conditioned generation across models of different sizes and accessibility levels. All models were evaluated under varying decoding temperatures within the range \(\{0, 0.2, 0.4, 0.6, 0.8, 1.0\}\), to examine how stochasticity affects response quality and semantic consistency.
\begin{table*}[!htbp]
    \caption{Performance comparison across LLMs with and without persona augmentation. \textbf{Bold} indicates improved scores with PARAN. Avg. $\Delta$ denotes the average improvement over six models.}
    \centering
    \begin{tabular}{l|cccc}
        \toprule
        \textbf{LLMs} & \textbf{Rouge-2} & \textbf{BLEU} & \textbf{METEOR} & \textbf{Distinct-2} \\
        \midrule
        GPT-4o mini              & 0.0553 & 0.0253 & 0.2312 & 0.0760 \\
        \rowcolor{gray!10}
        \quad + PARAN            & \textbf{0.0573} & \textbf{0.0281} & \textbf{0.2390} & \textbf{0.0802} \\
        \midrule
        GPT-3.5 Turbo            & 0.0578 & 0.0290 & 0.2172 & 0.0769 \\
        \rowcolor{gray!10}
        \quad + PARAN            & 0.0555 & 0.0270 & 0.2134 & 0.0738 \\
        \midrule
        Claude 3.5 Haiku         & 0.0349 & 0.0145 & 0.1910 & 0.0766 \\
        \rowcolor{gray!10}
        \quad + PARAN            & \textbf{0.0439} & \textbf{0.0199} & \textbf{0.2184} & \textbf{0.0959} \\
        \midrule
        Llama 3.1 Instruct (8B)  & 0.0914 & 0.0485 & 0.2841 & 0.0929 \\
        \rowcolor{gray!10}
        \quad + PARAN            & \textbf{0.1081} & \textbf{0.0607} & \textbf{0.3061} & \textbf{0.0930} \\
        \midrule
        Llama 3.1 Instruct (70B) & 0.0696 & 0.0360 & 0.2498 & 0.0964 \\
        \rowcolor{gray!10}
        \quad + PARAN            & 0.0631 & 0.0334 & 0.2345 & 0.0808 \\
        \midrule
        Amazon Nova Lite         & 0.0540 & 0.0268 & 0.2292 & 0.0795 \\
        \rowcolor{gray!10}
        \quad + PARAN            & \textbf{0.0587} & \textbf{0.0289} & \textbf{0.2400} & 0.0778 \\
        \midrule
        \textbf{Avg. $\Delta$}   & 7.76  & 11.20 & 3.72  & 1.41  \\
        \bottomrule
    \end{tabular}
    \label{tab:main_results}
\end{table*}

\section{Experiments}

In this section, we conduct experiments to address the following research questions:
\begin{itemize}
    \item \textbf{RQ1}: Can LLMs generate responses that preserve review content while reflecting personalized variation?
    \item \textbf{RQ2}: How effectively can LLMs simulate personas when the temperature of LLM variables varies?
    \item \textbf{RQ3}: How do the individual components of the framework contribute to its overall performance?
    % \item \textbf{RQ4}:
\end{itemize}

\subsection{Overall Performance (RQ1)}

To evaluate the effectiveness of our proposed PARAN framework—which incorporates both explicit and implicit persona representations—we conducted a comparative analysis across six different LLMs. Table II presents the results across four evaluation metrics: Rouge-2, BLEU, METEOR, and Distinct-2.

We observed that the integration of PARAN yielded notable performance improvements over the baseline setting without any persona conditioning in most models, though the magnitude of improvement varied significantly depending on the model architecture and (in the case of open models) parameter size.

\noindent \textbf{Performance Gains by Model Type}.
Models such as GPT-4o mini, Claude 3.5 Haiku, and Llama 3.1 Instruct (8B) demonstrated clear and consistent improvements across all metrics with the inclusion of PARAN. For instance, Claude 3.5 Haiku improved by 25.79\% in Rouge-2, 37.24\% in BLEU, and 25.2\% in Distinct-2, suggesting that persona conditioning can substantially benefit smaller or instruction-tuned models with moderate capacity.

In contrast, GPT-3.5 Turbo and Llama 3.1 Instruct (70B) showed marginal or even negative changes upon adding PARAN. This suggests that very large models may already possess sufficient contextual and stylistic flexibility, rendering additional persona signals either redundant or, in some cases, even distracting.

\noindent \textbf{Effect of Model Size within the Llama Series}.
The Llama series provides a particularly interesting case study due to the public availability of model size and architectural details. Specifically, Llama 3.1 (8B) benefits significantly from the proposed PARAN framework, achieving substantial gains of +18.27\% in Rouge-2, +25.15\% in BLEU, +7.74\% in METEOR, and +0.11\% in Distinct-2. In contrast, Llama 3.1 (70B) exhibits declines or negligible gains across the same metrics, with Rouge-2, BLEU, METEOR, and Distinct-2 decreasing by -9.34\%, -7.22\%, -6.12\%, and -16.18\% respectively.

This divergence highlights an important phenomenon: smaller models appear to benefit more from persona-based guidance, possibly because they lack the capacity to implicitly model nuanced user-specific preferences. In contrast, larger models may already encode such diversity internally, and additional persona conditioning may offer limited or even adverse effects. This aligns with findings \cite{fewshot} from few-shot learning literature, where smaller models require stronger external signals, while larger models can generalize with minimal prompting.

\begin{figure}[!htbp]
  \centering
  \begin{subfigure}[t]{0.235\textwidth}
    \centering
    \includegraphics[width=\linewidth]{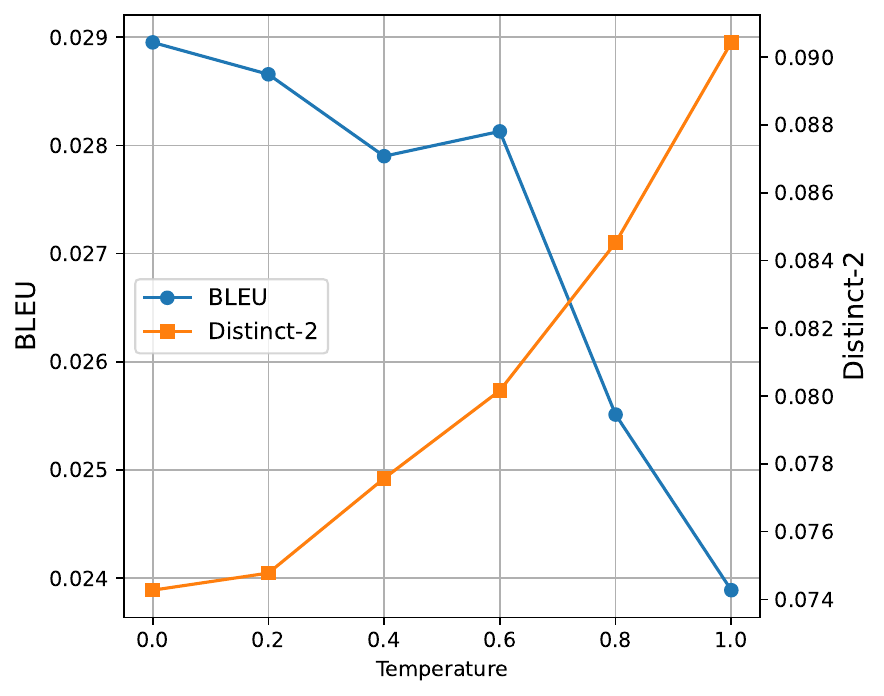}
    \caption{GPT-4o mini}
    \label{fig:gpt4o_tradeoff}
  \end{subfigure}
  \hfill
  \begin{subfigure}[t]{0.235\textwidth}
    \centering
    \includegraphics[width=\linewidth]{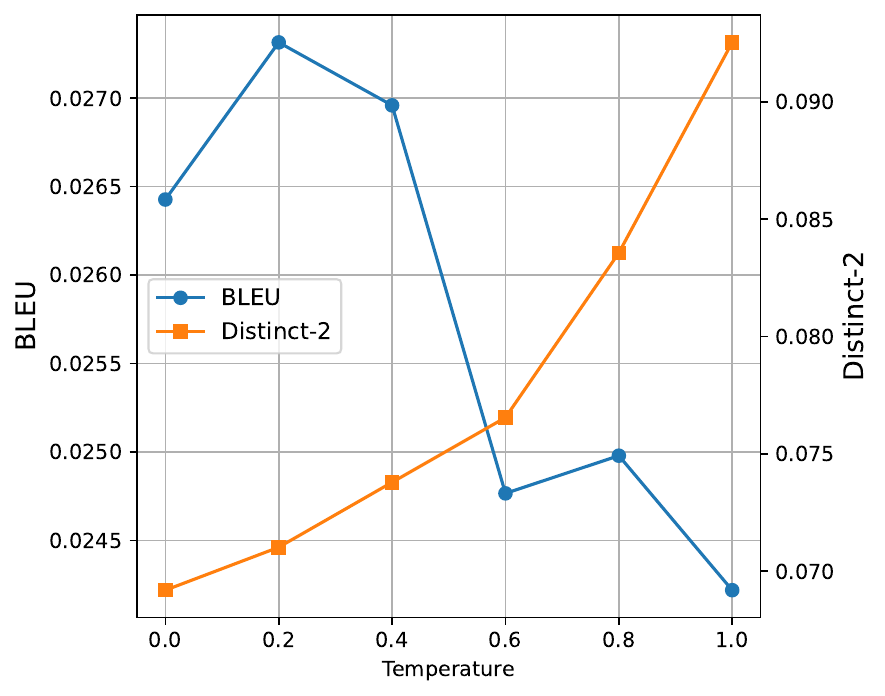}
    \caption{GPT-3.5 Turbo}
    \label{fig:gpt4o_additional}
  \end{subfigure}

  \vspace{0.5cm}

  \begin{subfigure}[t]{0.235\textwidth}
    \centering
    \includegraphics[width=\linewidth]{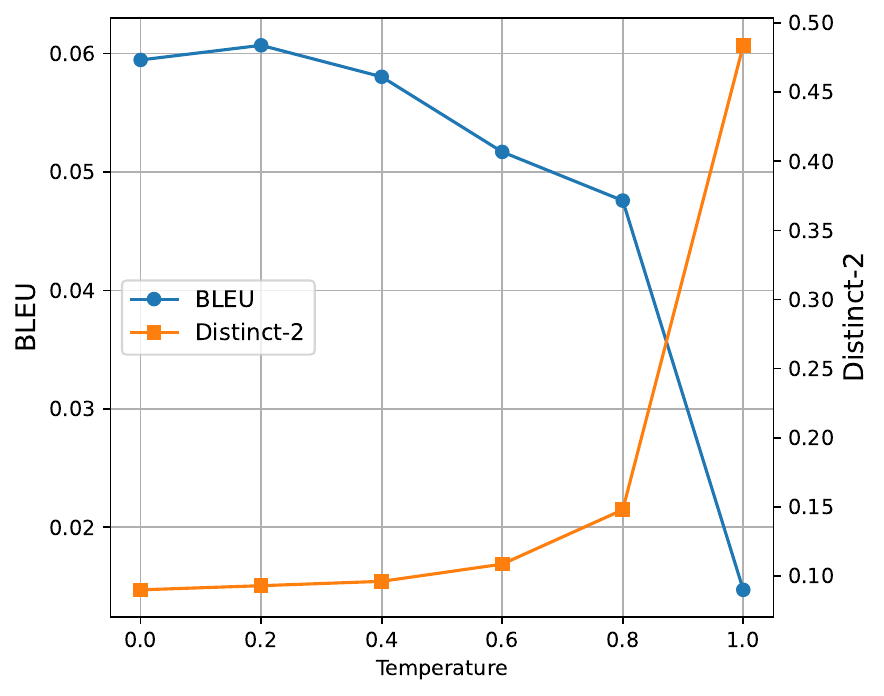}
    \caption{Llama 3.1 Instruct (8B)}
    \label{fig:llama8b_tradeoff}
  \end{subfigure}
  \hfill
  \begin{subfigure}[t]{0.235\textwidth}
    \centering
    \includegraphics[width=\linewidth]{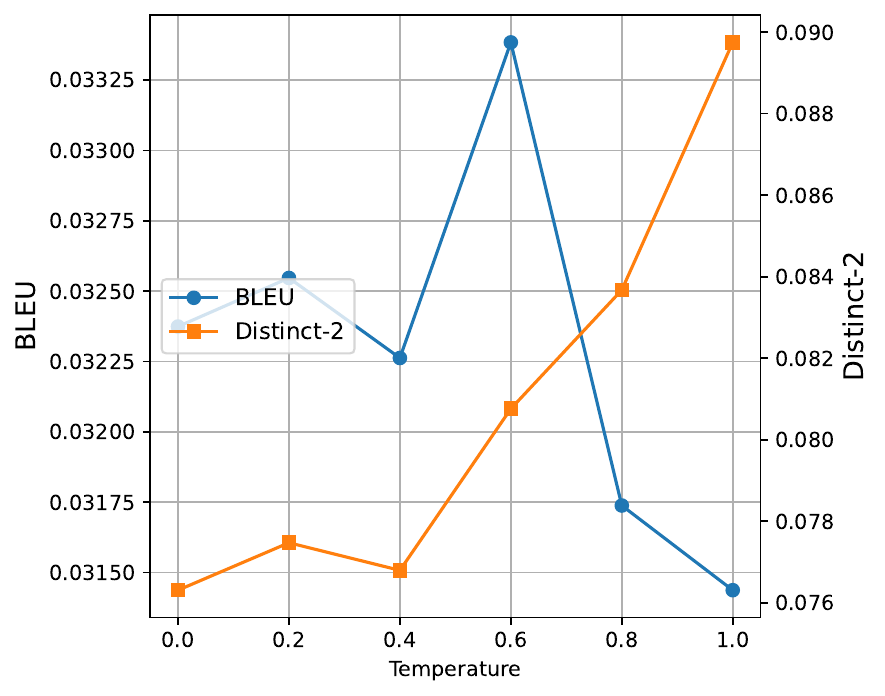}
    \caption{Llama 3.1 Instruct (70B)}
    \label{fig:llama8b_additional}
  \end{subfigure}

  \caption{Trade-off between lexical precision and response diversity across temperature settings}
  \label{fig:tradeoff_matrix}
\end{figure}

\noindent \textbf{Interpretation and Implications}.
These results suggest that the effectiveness of persona integration is model-dependent. For compact or medium-sized LLMs, explicit persona signals may act as informative inductive biases, improving alignment and specificity. However, for larger models, especially those in the 70B+ range, additional persona layers may lead to performance saturation or even overconditioning.

% version for author name
% \begin{figure}[!htbp]
%   \centering
%   \begin{subfigure}[t]{0.235\textwidth}
%     \centering
%     \includegraphics[width=\linewidth]{fig/hyperparams_gpt-4o-mini.pdf}
%     \caption{GPT-4o mini}
%     \label{fig:gpt4o_tradeoff}
%   \end{subfigure}
%   \hfill
%   \begin{subfigure}[t]{0.235\textwidth}
%     \centering
%     \includegraphics[width=\linewidth]{fig/hyperparams_gpt-3.5-turbo.pdf}
%     \caption{GPT-3.5 Turbo}
%     \label{fig:gpt4o_additional}
%   \end{subfigure}

%   \vspace{0.5cm}

%   \begin{subfigure}[t]{0.235\textwidth}
%     \centering
%     \includegraphics[width=\linewidth]{fig/hyperparams_llama3-1-8b.pdf}
%     \caption{Llama 3.1 Instruct 8B}
%     \label{fig:llama8b_tradeoff}
%   \end{subfigure}
%   \hfill
%   \begin{subfigure}[t]{0.235\textwidth}
%     \centering
%     \includegraphics[width=\linewidth]{fig/hyperparams_llama3-1-70b.pdf}
%     \caption{Llama 3.1 Instruct 70B)}
%     \label{fig:llama8b_additional}
%   \end{subfigure}

%   \caption{Trade-off between lexical precision and response diversity across temperature settings}
%   \label{fig:tradeoff_matrix}
% \end{figure}

\subsection{Hyperparameter Analysis (RQ2)}

% \begin{figure}[!htbp]
%   \centering
%   \begin{subfigure}[t]{0.235\textwidth}
%     \centering
%     \includegraphics[width=\linewidth]{fig/hyperparams_gpt-4o-mini.pdf}
%     \caption{GPT-4o mini}
%     \label{fig:gpt4o_tradeoff}
%   \end{subfigure}
%   \hfill
%   \begin{subfigure}[t]{0.235\textwidth}
%     \centering
%     \includegraphics[width=\linewidth]{fig/hyperparams_gpt-3.5-turbo.pdf}
%     \caption{GPT-3.5 Turbo}
%     \label{fig:gpt4o_additional}
%   \end{subfigure}

%   \vspace{0.5cm}

%   \begin{subfigure}[t]{0.235\textwidth}
%     \centering
%     \includegraphics[width=\linewidth]{fig/hyperparams_llama3-1-8b.pdf}
%     \caption{Llama 3.1 Instruct 8B}
%     \label{fig:llama8b_tradeoff}
%   \end{subfigure}
%   \hfill
%   \begin{subfigure}[t]{0.235\textwidth}
%     \centering
%     \includegraphics[width=\linewidth]{fig/hyperparams_llama3-1-70b.pdf}
%     \caption{Llama 3.1 Instruct 70B)}
%     \label{fig:llama8b_additional}
%   \end{subfigure}

%   \caption{Trade-off between lexical precision and response diversity across temperature settings}
%   \label{fig:tradeoff_matrix}
% \end{figure}

\noindent \textbf{Tradeoff between Precision and Diversity}.
Since we did not train to optimize a specific metric, we had to find the optimal case with appropriate criteria for each different metric evaluated. In this regard, we computed a sum-of-ranks \cite{sumofrank} score for each method. We ranked the methods each by Rouge-2, BLEU, METEOR, and Distinct-2, and then took a weighted sum of the four ranks, with 50\% of the weight assigned to Distinct-2, and 50\% distributed evenly among the other metrics. Based on this analysis, we selected the optimal temperature for each model as follows: GPT-4o mini (0.6), GPT-3.5 Turbo (0.4), Claude 3.5 Haiku (1.0), Llama 3.1 (8B) (0.2), Llama 3.1 (70B) (0.6), and Amazon Nova Lite (0.8). All main results in Table \ref{tab:main_results} are reported using these settings.

Figure \ref{fig:tradeoff_matrix} illustrates the trade-off dynamics between precision (measured by BLEU) and diversity (measured by Distinct-2) as temperature increases. While we evaluated all six models, we visualize four representative examples in Figure~\ref{fig:tradeoff_matrix}. While some models exhibit occasional fluctuations at specific temperature values, all four models overall follow the expected trade-off pattern: lexical precision (BLEU) decreases as temperature increases, while diversity (Distinct-2) rises accordingly. This suggests that decoding stochasticity consistently promotes more diverse outputs at the expense of surface-level accuracy across different LLMs.

\noindent \textbf{Semantic Robustness to Temperature Variation}. 
To evaluate the consistency of model outputs under different decoding stochasticity, we analyze the BERTScore F1 \cite{2020BERTScore:} between the original review and the generated responses across a range of temperature values (0.0 to 1.0). Unlike token-level precision or diversity metrics, which are highly sensitive to surface-level variation, BERTScore leverages contextual embeddings to assess semantic alignment, making it suitable for evaluating semantic robustness across variations.
% version for author name
\begin{figure}[!htbp]
  \centering
  \includegraphics[width = 3.4in]{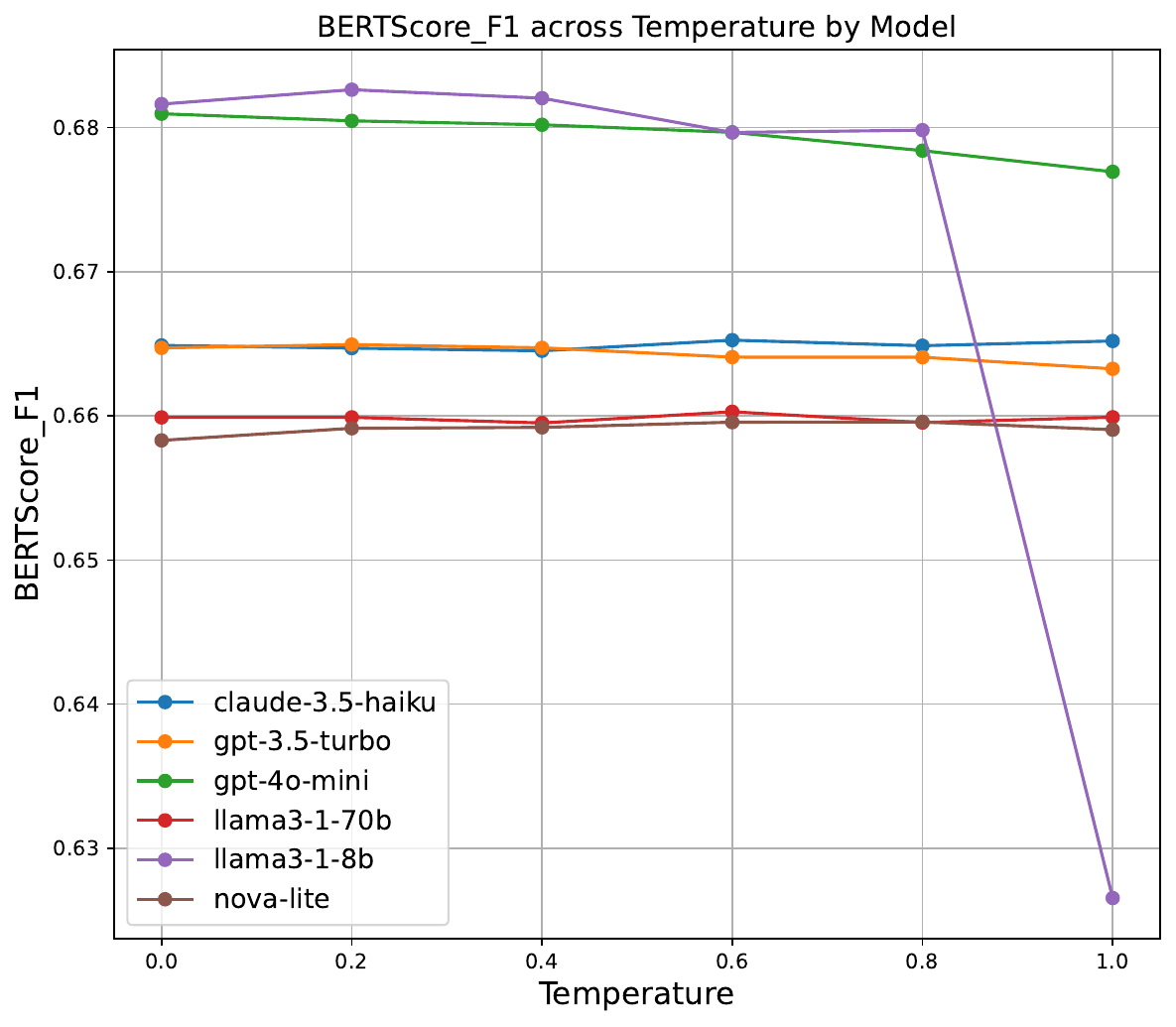}
  \caption{Semantic Robustness Across Temperature Settings.}
  \label{fig:answer_consistency}
\end{figure}

As shown in Figure \ref{fig:answer_consistency}, the BERTScore F1 of most models remains remarkably stable despite increasing temperature, indicating that semantic consistency is largely preserved even when the model is prompted to generate more diverse or creative responses. This robustness suggests that the core message of the review is faithfully maintained across generations, regardless of surface-level changes in expression.

% Notably, GPT-4o-mini and GPT-3.5-Turbo show flat performance curves, while Claude-3.5-Haiku and Nova Lite exhibit only marginal fluctuations. This indicates that their decoding process consistently preserves semantic meaning across sampling variations.

Overall, LLMs exhibit a relatively flat performance curve across temperature values, indicating strong resilience to stochastic decoding. Among them, GPT-4o mini and Llama 3.1 Instruct (8B) consistently show the highest BERTScore F1, with only slight decreases as temperature increases. Notably, Llama 3.1 Instruct (8B) experiences a significant drop in BERTScore F1 at temperature 1.0. We hypothesize that this decline stems from overgeneration or semantic drift at high temperature settings, where the model may produce responses that are syntactically fluent but semantically less grounded in the input review. This suggests that Llama 8B’s sensitivity to sampling temperature is higher compared to other models, potentially due to architectural or training-related factors that affect generation stability.

These findings affirm that LLMs, when sufficiently capable, can maintain answer consistency across stochastic decoding settings, reinforcing the reliability of persona-based generation frameworks under practical usage conditions.

% \begin{figure}[!htbp]
%   \centering
%   \includegraphics[width = 3.4in]{fig/answer_consistency.pdf}
%   \caption{Semantic Robustness Across Temperature Settings.}
%   \label{fig:answer_consistency}
% \end{figure}

\subsection{Ablation Study (RQ3)}

\begin{table*}[!htbp]
    \caption{Ablation study comparing the effects of explicit and implicit persona components across different LLMs.}
    \centering
    \scalebox{0.85}{
    \begin{tabular}{l|l|cccc|l|cccc}
        \toprule
        \textbf{LLMs} & \textbf{Method} & \textbf{Rouge-2} & \textbf{BLEU} & \textbf{METEOR} & \textbf{Distinct-2} 
        & \textbf{LLMs} & \textbf{Rouge-2} & \textbf{BLEU} & \textbf{METEOR} & \textbf{Distinct-2} \\
        \midrule
        \multirow{5}{*}{GPT-4o mini} 
        & \textbf{PARAN}              & \underline{0.0573} & \underline{0.0281} & \underline{0.2390} & \textbf{0.0802} 
        & \multirow{5}{*}{Llama 3.1 Instruct (8B)} 
        & \textbf{0.1081} & \textbf{0.0607} & \textbf{0.3061} & \underline{0.0930} \\
        & w Explicit Persona & \textbf{0.0601} & \textbf{0.0302} & \textbf{0.2431} & 0.0774  
        & & \underline{0.1046} & \underline{0.0596} & \underline{0.3027} & \textbf{0.0938} \\
        & w Implicit Persona & 0.0514 & 0.0243 & 0.2289 & \underline{0.0778} 
        & & 0.0895 & 0.0497 & 0.2744 & 0.0879 \\
        & w/o E\&I Persona   & 0.0535 & 0.0253 & 0.2312 & 0.0760 
        & & 0.0914 & 0.0485 & 0.2841 & 0.0929 \\
        & $\Delta$           & 7.10 & 11.07 & 3.37 & 5.53 
        & & 18.27  & 25.15  & 7.74   & 0.11 \\
        \midrule
        \multirow{5}{*}{GPT-3.5 Turbo} 
        & \textbf{PARAN}              & 0.0555 & 0.0270 & 0.2134 & 0.0738 
        & \multirow{5}{*}{Llama 3.1 Instruct (70B)} 
        & 0.0631 & 0.0334 & 0.2345 & 0.0808 \\
        & w Explicit Persona & \textbf{0.0657} & \textbf{0.0332} & \textbf{0.2306} & \textbf{0.0789}  
        & & \underline{0.0685} & \textbf{0.0360} & \underline{0.2455} & \underline{0.0827} \\
        & w Implicit Persona & 0.0554 & 0.0255 & \underline{0.2191} & \underline{0.0786} 
        & & 0.0543 & 0.0274 & 0.2209 & 0.0822 \\
        & w/o E\&I Persona   & \underline{0.0578} & \underline{0.0290} & 0.2172 & 0.0769 
        & & \textbf{0.0696} & \textbf{0.0360} & \textbf{0.2498} & \textbf{0.0964} \\
        & $\Delta$           & -3.98 & -6.9 & -1.75 & -4.03 
        & & -9.34  & -7.22  & -6.12  & -16.18 \\
        \midrule
        \multirow{5}{*}{Claude 3.5 Haiku} 
        & \textbf{PARAN}              & \textbf{0.0439} & \textbf{0.0199} & \textbf{0.2184} & \underline{0.0959}  
        & \multirow{5}{*}{Amazon Nova Lite} 
        & \textbf{0.0587} & \textbf{0.0289} & \textbf{0.2400} & 0.0778 \\
        & w Explicit Persona & \underline{0.0437} & \underline{0.0192} & 0.2160 & 0.0952 
        & & \underline{0.0586} & \underline{0.0290} & \underline{0.2389} & 0.0788 \\
        & w Implicit Persona & 0.0424 & 0.0191 & \underline{0.2163} & \textbf{0.1041}
        & & 0.0511 & 0.0240 & 0.2263 & \textbf{0.0815} \\
        & w/o E\&I Persona   & 0.0349 & 0.0145 & 0.1910 & 0.0766
        & & 0.0540 & 0.0268 & 0.2292 & \underline{0.0795} \\
        & $\Delta$           & 25.79 & 37.24 & 14.35 & 25.2
        & & 8.7   & 7.84   & 4.71   & -2.14 \\
        \bottomrule
    \end{tabular}}
    \label{tab:ablation}
\end{table*}

To understand the individual and combined contributions of the explicit and implicit persona modules, we conducted an ablation study as shown in Table \ref{tab:ablation} across six different LLMs. Each model was tested under four settings: with both persona types (PARAN), with only the explicit persona (w Explicit Persona), with only the implicit persona (w Implicit Persona), and with neither (w/o E\&I Persona). The final row ($\Delta$) quantifies the change in performance when using PARAN relative to the baseline without persona conditioning (w/o E\&I Persona). This setup allows us to assess (1) whether each module effectively enhances its intended dimension—precision or diversity—and (2) whether the combination of both yields synergistic or conflicting effects. 

% \textbf{Targeted Impact of Each Persona Module}
\noindent \textbf{Effect of Explicit Persona}.
Explicit Persona aims to improve precision-oriented metrics—Rouge-2, BLEU, and METEOR—by grounding the generation in factual content. Across nearly all models, adding only the explicit persona (w Explicit Persona) led to higher precision scores than the baseline (w/o E\&I Persona). Notably, GPT-3.5 Turbo shows a significant improvement with only the explicit persona: +13.67\% in Rouge-2, +14.48\% in BLEU, and +6.17\% in METEOR. Claude 3.5 Haiku also demonstrates strong gains in precision-oriented metrics, with a 25.21\% increase in Rouge-2, a 32.41\% increase in BLEU, and a 13.09\% increase in METEOR compared to the baseline.

These results confirm that explicit persona information effectively enhances the relevance and fidelity of generated responses.

\noindent \textbf{Effect of Implicit Persona}.
Implicit Persona, by contrast, targets diversity enhancement via stylistic and user-specific variation. This is best reflected in the Distinct-2 metric. Claude 3.5 Haiku achieves the highest Distinct-2 score (0.1041) when using only the implicit persona, a +35.9\% gain over the baseline. GPT-4o mini, GPT-3.5 Turbo, and Amazon Nova Lite also exhibit moderate increases in Distinct-2 of 2.37\%, 2.21\%, and 2.52\%, respectively, when only the implicit persona is used.

This supports the hypothesis that LLMs can use latent persona signals to diversify their responses without explicit conditioning.

\noindent \textbf{Combined Effects of PARAN: Trade-off vs. Synergy}.
While combining both persona modules (PARAN) generally improves overall performance across most metrics, the joint effect sometimes dampens individual gains due to a trade-off between precision and diversity.

For instance, GPT-4o mini achieves the best precision metrics (Rouge-2, BLEU, METEOR) when only the explicit persona is used, yet its PARAN variant remains second-best in precision while achieving the highest Distinct-2 (0.0802) among all variants. This indicates that PARAN maintains a favorable balance across objectives, achieving strong diversity without substantially compromising precision. Similarly, Llama 3.1 (8B) with PARAN outperforms all other methods across Rouge-2, BLEU, and METEOR, while maintaining competitive Distinct-2. This suggests that certain models, particularly mid-sized ones, can absorb both persona types without suffering trade-offs.

\noindent \textbf{Degradation under Combined Persona}.
Interestingly, GPT-3.5 Turbo shows degraded performance with PARAN compared to using only the explicit persona: explicit persona alone yields the highest Rouge-2 (0.0657), BLEU (0.0332), and METEOR (0.2306), while PARAN leads to reductions in all three. This suggests interference or over-conditioning, where the presence of implicit persona information may have introduced stylistic shifts that detract from factual alignment.

Llama 3.1 (70B) exhibits an even more striking case: the best performance across precision and diversity metrics is observed when no persona conditioning is applied at all (w/o E\&I Persona), outperforming both explicit/implicit-only and PARAN settings. This result implies that large-scale models with strong pretrained contextual and reasoning capabilities may already encode sufficient user-aligned variability internally. Introducing external persona signals in such cases might constrain or distort the generation space, leading to degraded alignment and fluency. In other words, for highly capable LLMs, additional conditioning may not only be redundant but potentially detrimental when not carefully calibrated.

\noindent \textbf{Selective Persona Conditioning for Targeted Outcomes}.
The ablation results show that explicit and implicit persona modules benefit different aspects of response generation. Explicit persona improves precision-focused metrics (e.g., ROUGE-2, BLEU, METEOR) by grounding responses in review content. Implicit persona enhances diversity (e.g., Distinct-2) through personalized variation inferred from user traits. These effects are consistent across most LLMs tested.

Depending on the desired outcome—factual fidelity or personalized diversity—one can selectively apply the appropriate persona module. The combined PARAN framework often balances both, though minor trade-offs may occur. In some cases (e.g., GPT-3.5 Turbo), using only explicit persona even outperformed the full setup. This finding suggests that persona selection should be tailored to each model. Overall, persona conditioning in LLMs can be modular, adaptive, and goal-driven.
\section{Conclusion}

We introduced PARAN, a prompting-based framework for generating personalized and contextually grounded responses to food delivery reviews. By combining explicit personas derived from review content with implicit personas inferred from linguistic and stylistic cues, PARAN effectively balances content fidelity to review content with stylistic diversity. Experimental results across six LLMs demonstrate the modular benefits of each persona type and reveal how temperature influences sensitivity to persona conditioning. Our analysis highlights that while explicit personas improve precision, implicit personas contribute to diversity, and their joint use often leads to favorable trade-offs. Importantly, our approach requires no fine-tuning, making it lightweight and easily deployable. 

PARAN’s lightweight design makes it applicable to real-world settings such as semi-automated customer engagement on food-delivery platforms. As next steps, we will (i) move beyond diversity as an indirect proxy by developing direct personalization metrics; (ii) quantitatively compare PARAN against fine-tuned approaches to identify the regimes where persona-prompting is preferable versus fine-tuning; and (iii) examine the robustness of persona inference under noisy or adversarial inputs.

\bibliographystyle{IEEEtran}
\bibliography{references}

\end{document}